\documentclass[conference]{IEEEtran}

\usepackage{amsmath,amssymb,amsfonts}
\usepackage{algorithmic}
\usepackage{algorithm}
\usepackage{graphicx}
\usepackage{textcomp}
\usepackage{booktabs}
\usepackage{array}
\usepackage{scrextend}
\usepackage[normalem]{ulem}
\usepackage{subcaption}
\usepackage{xcolor}
\usepackage{tabularx}
\usepackage{orcidlink}
\usepackage{ifpdf}
\usepackage{caption}
\usepackage[backend=biber, url=false, hyperref, natbib, sorting=none, style=numeric-comp]{biblatex}

\def\BibTeX{{\rm B\kern-.05em{\sc i\kern-.025em b}\kern-.08em
    T\kern-.1667em\lower.7ex\hbox{E}\kern-.125emX}}

\makeatletter
\newcommand{\linebreakand}{%
  \end{@IEEEauthorhalign}
  \hfill\mbox{}\\
  \mbox{}\hfill\begin{@IEEEauthorhalign}
}
\makeatother

\begin{document}

\title{Improving Applicability of Deep Learning based Token Classification models during Training}

\author{
\IEEEauthorblockN{Anket Mehra, Malte Prieß\orcidlink{0009-0004-4626-2513}}
\IEEEauthorblockA{\textit{Department of Computer Science and Electrical Engineering} \\
\textit{University of Applied Sciences Kiel}\\
Kiel, Germany \\
\tt anket.mehra@student.fh-kiel.de\,|\,malte.priess@fh-kiel.de}
\linebreakand \\
\IEEEauthorblockN{Marian Himstedt}
\IEEEauthorblockA{\textit{Department of Electrical Engineering and Computer Science} \\
\textit{Technical University of Applied Sciences Lübeck}\\
Lübeck, Germany \\
\tt marian.himstedt@th-luebeck.de}
}

\ifpdf
\pdfoutput=1 
\pdfcompresslevel=9     
\pdftrue
\pdfinfo{
	/Author     (Anket Mehra; Malte Prie{\ss}; Marian Himstedt)
	/Title      (Improving Applicability of Deep Learning based Token Classification models during Training)
	/Subject    (Deep Learning Token Classification)
	/Keywords   (document analysis; natural language processing; token classification; business process automation; deep learning; model training)
}
\pdfminorversion=5
\fi

\makeatletter
\def\mycopyrightnotice{
	{\footnotesize
		\begin{minipage}{0.8\textwidth}
			\centering
			Please cite as: XXX.}
		\end{minipage}
}

\makeatother

\maketitle

\begin{abstract}
This paper shows that further evaluation metrics during model training are needed to decide about its applicability in inference. As an example, a LayoutLM-based model is trained for token classification in documents. The documents are German receipts. We show that conventional classification metrics, represented by the F1-Score in our experiments, are insufficient for evaluating the applicability of machine learning models in practice. To address this problem, we introduce a novel metric, Document Integrity Precision (DIP), as a solution for visual document understanding and the token classification task. To the best of our knowledge, nothing comparable has been introduced in this context. DIP is a rigorous metric, describing how many documents of the test dataset require manual interventions. It enables AI researchers and software developers to conduct an in-depth investigation of the level of process automation in business software. In order to validate DIP, we conduct experiments with our created models to highlight and analyze the impact and relevance of DIP to evaluate if the model should be deployed or not in different training settings. Our results demonstrate that existing metrics barely change for isolated model impairments, whereas DIP indicates that the model requires substantial human interventions in deployment. The larger the set of entities being predicted, the less sensitive conventional metrics are, entailing poor automation quality. DIP, in contrast, remains a single value to be interpreted for entire entity sets. This highlights the importance of having metrics that focus on the business task for model training in production. Since DIP is created for the token classification task, more research is needed to find suitable metrics for other training tasks.
\end{abstract}


\section{Introduction}

As businesses globally move toward digital transformation, the efficiency of processing financial documents such as invoices has become a key point of operational optimization. Invoice recognition automation not only reduces manual labor but also minimizes errors, accelerates transaction cycles, and enhances data analytics capabilities \citep{Cai2023, Kumar2023, Tuan2021, Gunaratne2021}.

In the field of digital document processing, the introduction of deep learning models has opened up a new level of efficiency and accuracy, including in the domain of invoice recognition \citep{Xu2020, Xu2022, Huang2022, donut2021, trie2021}.

Achieving a trade-off between cutting-edge performance and accessibility, licensing constraints, and model adaptability is crucial in a commercial setting. However, finding the suitable model for business applications is not simple, since new document analysis systems, capable of performing classifications at the token level, are only evaluated by conventional classification metrics such as precision, recall, and F1-Score \citep{donut2021, LayoutLM, Geron2022}, missing information about the model's performance in inference and ignoring the aspect of applicability. A mismatch between research and applicability for model training in practice is exposed, as the experiments in this paper show. Here, the applicability of a model is defined by how often its output needs to be manually corrected in a post-processing step. More evaluation metrics are needed that focus on the applicability aspect of a model for business environments. To address this problem, a new metric for the token classification task in visual document understanding is introduced, called Document Integrity Precision (DIP). It will be explained in more detail after the introduction of our research questions and in Section \ref{sec:methodology}.        
To understand DIP's importance, the following research question is examined: 
\begin{itemize}
    \item[RQ:] Why are further evaluation metrics important in evaluating the applicability of a deep learning model?   
\end{itemize}
\begin{figure*}
    \centering
    \includegraphics[width=12cm]{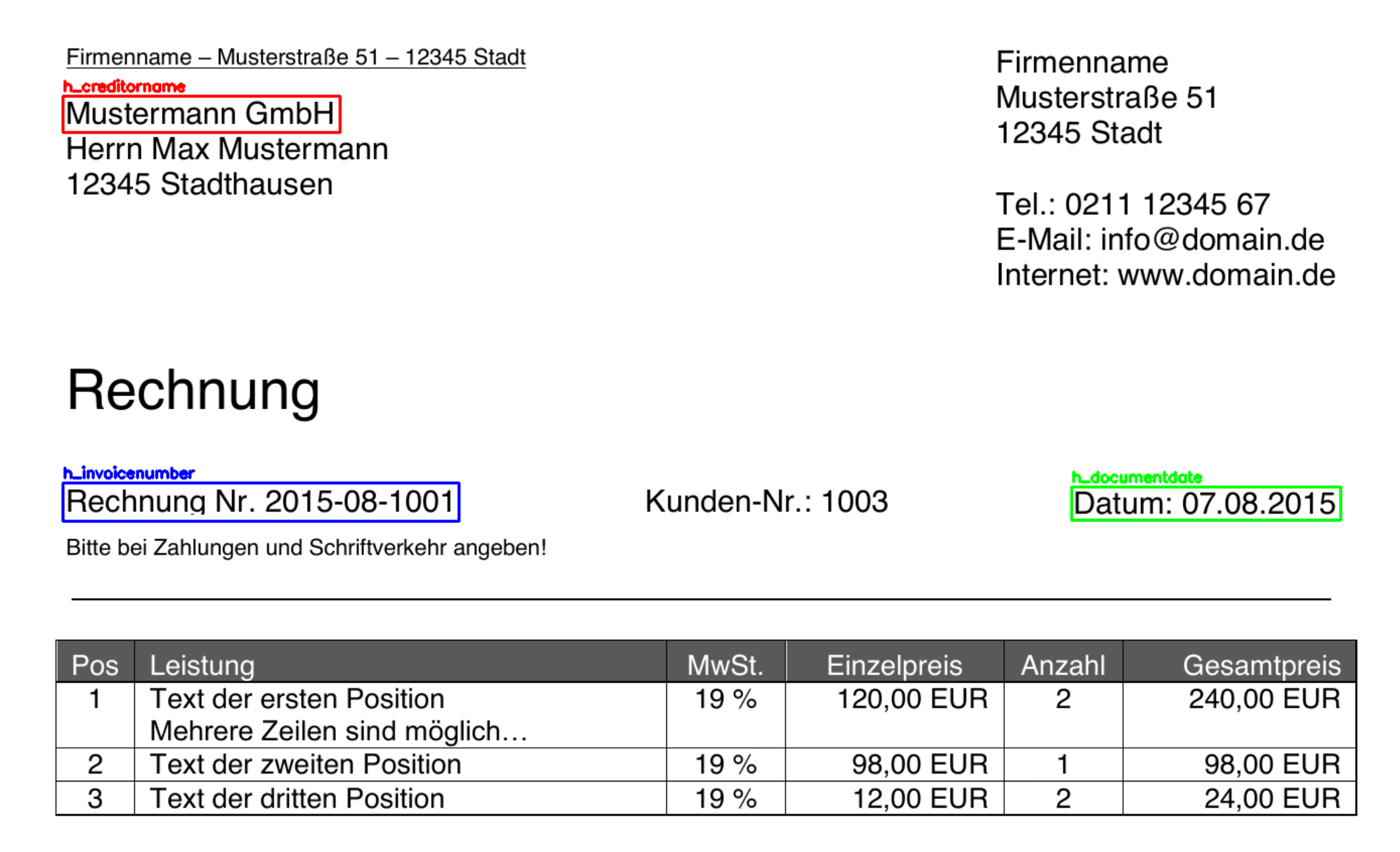}
    \caption{Visualized exemplary output of a token classification model. The boxes indicate tokens with labels of interest such as the name of the creditor (red), the invoice number (blue) or the invoice date (green). These tokens could be used for data-analysis or further processing, like automatically filling out archive-systems or web-forms.} 
    \label{fig:tcOutput}
\end{figure*}
To answer the research question, exemplary deep learning models for token classification are trained. We use a dataset of German receipts. For the training, two scenarios for splitting the data were chosen to mimic two potential inference environments while the model is used in production. For each model, the F1-Score, as a representative of conventional training metrics for classification, and DIP are calculated. The results will be compared and analyzed.

The models are based on the architecture of LayoutLMv1 because of its ease of use with the transformers library \citep{wolf2020huggingfacestransformersstateoftheartnatural}. \textit{We train a foundational version to showcase that conventional metrics such as the F1-Score are not suitable to justify whether a model should be used in inference or not. Adding further models, maybe newer, and their results will not assist in novelty and relevance, since the problems addressed are fundamental with respect to conventional metrics and therefore independent of the exact architecture of the deep learning model used}.

An exemplary output of such a model in production is visualized in Figure \ref{fig:tcOutput}. To add the aspect of applicability, as mentioned above, DIP is also calculated and shows that an optimally trained model -- indicated by high conventional metric scores (above 90\%) -- does not necessarily mean it is suitable for use in production. DIP incorporates the number of documents whose tokens are correctly classified in their entirety with respect to all classified documents. It addresses the need for a high degree of automation. This metric is of major importance from a business perspective, as it directly relates to the operational efficiency of the automated invoice processing system. However, it does so with a focus on the Token Classification Task (TCT) for Visual Document Understanding (VDU). Further metrics for other training goals and business cases need to be created, such as for regression or prediction tasks.

The remainder of the paper has the following structure: In Section \ref{sec:RelWork}, we briefly review state-of-the-art models for document analysis suitable for invoice recognition and give a brief introduction to the domain of information extraction. In Section \ref{sec:methodology}, we dive into more detail about the methodology, including a short description of the model's architecture, the required component of Optical Character Recognition (OCR), the preparation of the applied ground truth data, as well as measures used to evaluate the model's performance and suitability for usage in inference. Section \ref{sec:trainSetup} includes a description of the training setup, covering the utilized hyperparameters, the selected labels of the classification task, and the underlying hardware setup. Numerical results and a discussion of relevant training runs are provided in Sections \ref{sec:res} and \ref{sec:discussion}. Section \ref{sec:concl} concludes with key findings and a brief outlook on future work.

\section{Information extraction and recent model-architectures} \label{sec:RelWork}
In this Section, we give a brief overview on information extraction on entities such as documents and how it evolved in the last years. Then, 2 deep learning model-architectures and their train process for information extraction are introduced in order to show by way of examples how these models are only trained via conventional metrics for classification.

\subsection{Information extraction}
\begin{table*}[htbp]
\centering
\caption{Documenttypes}
\label{tab:tableDocumentTypes}
\begin{tabularx}{\textwidth}{X|X|X}
\toprule
\textbf{Layout/Texttype} & \textbf{Structured} & \textbf{Semi-Structured} \\ 
\midrule
Fixed      & invoice, passport, ID card                        & business email  \\
\midrule
Variable              & purchase receipt, business card  & resume \\
\bottomrule
\end{tabularx}
\end{table*}

The extraction of information is currently highly discussed in research, especially with AI \citep{xu2024large, lu2022unified, zhu2023large, hong2022bros}. Before learning models became the standard, as f.e. LayoutLM or Donut, rule-based models were often used. Here, rules for pattern matching were defined with which key terms of documents \citep{sanderson2012} or linguistic features were extracted \citep{hobbs2010information}. To create the most useful patterns, domain expertise is needed. In addition, the problem of generalization on different documents remains, especially if the layout is unstructured \citep{lambruschini2024Trans}.
The first learning-based approaches for information extraction relied on analyzing only text data from a document \citep{ma-hovy-2016-end,yadav-bethard-2018-survey}. These models were coming from the area of natural language processing and tried to filter relevant key entities of the given textual context. Since documents, especially unstructured ones, contain visual information, the use of these multimodal models was introduced \citep{donut2021,LayoutLM,ConvulutionalLiu2021}.
Currently, enhancing multimodal retrieval with the optional usage of Large Language Models seems to be a point of focused interest as seen in \citep{ma2024unifying, li2024generative, long2024generative, wang2025cross}. Instead of only relying on textual features of tokens in documents, researchers added several document features to improve the extraction of key entities and try to achieve that the learning model is able to generalize, even on new documents. The already proposed models, LayoutLM and Donut, f.e. add position and layout information within their models to create richful embeddings for further processing within the model. Other approaches, as in \citep{ConvulutionalLiu2021} modeled their documents as graphs of text segments interconnected with each other through edges representing visual dependencies, e.g. distances between the text segments of the same document, to combine visual and textual data of a document. Another example is shown in \citep{ZhangNaep2024}, where the researchers leveraged the capabilities of generative artificial intelligence to extract information from an additional set of documents using the retrieval-augmented generation framework as in \citep{gao2024retrievalaugmentedgenerationlargelanguage}. 
In the domain of information extraction in documents, there is a differentiation between different document types such as the table \ref{tab:tableDocumentTypes} displays, analogous to \citep{trie2021}.

While there are several approaches for the information extraction task itself, research on the applicability of such methods and their training in enterprises is overseen. For models based on deep learning, there is no discussion of the correct train metric. Conventional metrics are used as is.
\subsection{LayoutLM}
For several tasks related to natural language processing, attention-based transformer models are used \citep{Vaswani2023, Mehra+Prieß+Peters}. However, their input is limited to be completely text-based. On the other hand, documents can be seen as multimodal entities, which do not only consist of text information but also visual metadata. Therefore, to leverage the multimodal aspects of a document, using a plain transformer model is not sufficient \citep{LayoutLM}.
LayoutLM is a deep learning model for document analysis and recognition. It is based on a bidirectional attention-based transformer model for processing text input. To leverage the model in analyzing documents, the input documents are expanded with information about the 2D-position of each token in a respective document. To further enhance the model on image-based tasks, an optional image embedding layer is added. For pre-training, a LayoutLM model is trained on the "Masked Visual-Language Model" (MVLM) task to learn a language while respecting the positions of each token in a document. In MVLM some input tokens are randomly hidden during pre-training and the model has to predict the correct token, while the positions are given \citep{LayoutLM}.
Optionally, to understand classes of documents, the model can be trained on the multi-label document classification task, learning to differ between several kinds of documents. Based on these pre-training tasks, LayoutLM gives a foundation to be fine-tuned \citep{LayoutLM}. 
On experiments done on several fine-tuning tasks, namely a form understanding, a receipt understanding, and an image classification task, the model has reached state of the art in each respective one. One downside performance-wise is that LayoutLM is not able to to capture information about text positions on it's own but rather depends on an upstream OCR to do so \citep{LayoutLM}.

Another important fact for the usage of LayoutLM is the licensing. The first version of the model is freely usable even in commercial settings because of its MIT license \citep{LayoutLMLicense}. Therefore, the entry barrier for using the model is reduced, compared to it's newer versions. Furthermore, it needs fewer data to train than Donut because it does not need to train a OCR-model but rather uses an external one to gather position information. Since the focus of this paper is the usage and applicability of the models and their train metrics, LayoutLMv1 - it's licensing corresponds to the ease of use and introduction in the software-landscape of enterprises hoped for and with it's capability of doing the TCT task, this paper will use a LayoutLM model for it's experiments.

\begin{algorithm*}
\caption{DIP algorithm}
\label{lst:dip}
\begin{algorithmic}
\REQUIRE predictedDocumentTokenLabels, trueDocumentTokenLabels
\ENSURE DIP

\STATE $amountWrongDocuments \gets 0$
\STATE $currentDocument \gets 0$
\FOR{each document in predictedDocumentTokenLabels}
    \STATE $currentToken \gets 0$
    \FOR{each predictedTokenLabel in document}
        \STATE $groundTruthLabel \gets trueDocumentTokenLabels[currentDocument][currentToken]$
        \IF{$predictedTokenLabel \neq groundTruthLabel$}
            \STATE $amountWrongDocuments \gets amountWrongDocuments + 1$
            \STATE \texttt{break}
        \ENDIF
        \STATE $currentToken \gets currentToken + 1$
    \ENDFOR
    \STATE $currentDocument \gets currentDocument 1$
\ENDFOR

\STATE $amountDocuments \gets len(predictedDocumentTokenLabels)$ \COMMENT{len(x) returns the length of an array}
\STATE $DIP \gets 1 - (amountWrongDocuments / amountDocuments)$
\RETURN $DIP$
\end{algorithmic}
\end{algorithm*}

\subsection{Donut}
Since the usage of OCR is performance-intensive and limits the token-analysis to one language, it seems sufficient to look for ways without using it. One OCR-less approach is Donut. Donut is an End-to-End transformer-based deep learning model, used for document analysis. At it's core, the model relies on an encoder-decoder architecture. The encoder receives as input the vector-representation of an image of a document, with a dimension of $H \times W \times 3$. $H$ is the document height, $W$ the respective width and 3 the respective RGB values of a pixel. The encoder, a swin transformer in the paper, receives the image, converts it into multiple patches, containing information about a part of the picture and creates an attention-based embedding for each patch. Afterwards, the embeddings of each patch run through 2 multi-layered perceptrons and then merged into one output $z$ \citep{donut2021}.
$z$ is used as input for the textual decoder. The decoder converts $z$ into a token sequence for the document, representing each token in order of appearance as one-hot vector of length $v$, with $v$ being the amount of words in the token-vocabulary used in the initial creation of the underlying BART-model \citep{donut2021}. Since it's an End-to-End model, there also is an output converter, which can convert the one-hot-vectors into a corresponding JSON format for each respective fine-tuning task.
To pre-train Donut, the authors used token prediction as binary classification task, in which the model learns to guess the correct words in an image. The decoder uses BART \citep{donut2021} as underlying transformer which is multilingual, making the model also multilingual.
In experiments, comparing the results of Donut with LayoutLM and newer versions in several downstream tasks, Donut showed promising results, by having higher scores in nearly all of them \citep{donut2021}.

\section{Methodology} \label{sec:methodology}

To answer the research question, we will train a deep learning model for token classification. The underlying model-architecture resembles LayoutLMv1. We've chosen this architecture because of its MIT license \citep{LayoutLMLicense} and ease of usage within the huggingface transformers library \citep{wolf2020huggingfacestransformersstateoftheartnatural}. We could have used other architectures such as Donut too. The only matter of concern for the choice of our model is that the underlying architecture allows it to be trained and later used for token classification. The reason is that DIP is not part of the model architecture but rather part of the training. It should be used as a loss metric or additionally to govern the model performance in the TCT. We've used DIP as additional train metric. The pseudo code for DIP as loss metric can be found in Listing \ref{lst:dip}. The inputs of the algorithm $predictedDocumentTokenLabels$ and $trueDocumentTokenLabels$ are arrays of documents. Each document consists either the prediction or ground truth data for the label of each token in the respective document. We trained the model with a private dataset of real German receipts. 

To generate the train data for LayoutLM, we had to get the positions of the tokens of interest in the document and the tokens themselves for serialization. Then, we automated the process of ground truth data creation. 

We've created two train scenarios S1 and S2. For each scenario a model was analyzed as mentioned above. In scenario S1 we ensure that the train and test data follow the same distribution of receipts of the same structure and creditor. This is done to mimic a optimal business environment in which the model is prepared for all potential receipts in inference. This scenario is flawed, as, we assume, most business environments are dynamic and creditors could change their receipt designs and new creditors with unknown receipt-structures could come over the time. This is the reason for S2. Here, the split between train and test data ensures that in the test data there are creditors unknown in the test data. Therefore, the model cannot train the structures of these receipts.            

For both scenarios, we trained in a fixed train environment, as listed in Section \ref{sec:trainSetup}, then calculated the F1-Score for each token class and DIP per epoch and compared the meaning of the results for the business task of TCT. 

\begin{figure*}[ht]
    \centering
    \includegraphics[width=\textwidth]{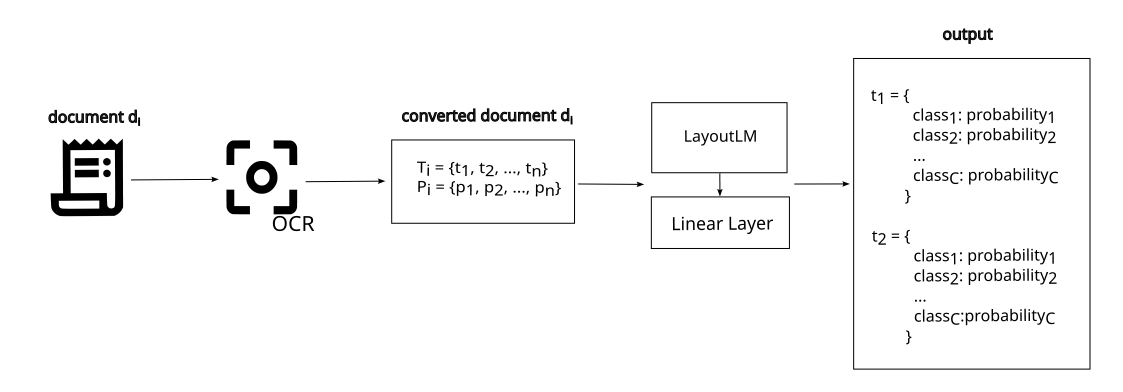}
    \caption{Overview of TCT system for a document $d_i$ with $n$ tokens and $c$ possible token classes}
    \label{fig:systemOverview}
\end{figure*}

\subsection{Task description} \label{subsec:taskDescr}
This paper fine-tunes a LayoutLMv1 model on TCT. For that, a document $d_i$ from a set of documents $D$ consists a set of tokens $T_i$ with their corresponding positions $P_i$ given by an OCR. If a document has $n$ tokens, the array $P_i$ has an dimension of $n \times 4$ because the position of the $n$-th token $t$ is represented by a vector $p_{n}$ as seen in \eqref{eq:p_1}.
\begin{equation} \label{eq:p_1}
    p_{n} = \{ x_1, y_1, x_2, y_2 \}
\end{equation}
$x_1$ and $y_1$ represent the upper left corner of a bounding box for each respective token while $x_2$ and $y_2$ represent the bottom right corner in pixels. Furthermore, each token has a class it can be mapped to. The finetuned LayoutLM model will take $d$ as input and outputs the probability for each class for each of the document's tokens from a pregiven set of classes $c$. The whole process is visualized in figure \ref{fig:systemOverview}, providing an overview of the created document analysis system for token classification task (TCT).

\subsection{OCR}
Optical character recognition (OCR) is a fundamental operation required by each visual document understanding (VDU) system. It enables the detection of text in visual documents which are subsequently processed by NLP systems. 
A multitude of open source software and commercial applications rely on the Tesseract engine \citep{tesseract} which employs image binarization, skew correction, line finding, blob detection and (handcrafted) feature description as well character recognition. Though recent engine versions include deep learning-based character recognition and text classification, Tesseract consistently performs worse than novel systems like EasyOCR \citep{easyOCR} omitting handcrafted feature extraction (see e.g. \citep{vedhaviyassh2022}). EasyOCR is an open source software provided by Jaded AI which includes CRAFT \citep{baek2019}, a deep learning-based text detection algorithm, and a Convolutional Recurrent Neural Network (CRNN) utilizing Convolutional and LSTM layers for text recognition \citep{shi2016}. Our approach makes use of EasyOCR which is equipped with a pre-trained model for German language. As a result of this step we obtain a set of tokens $T_i$ (OCR readings) for document $d_i$, with each token being supplemented by its image coordinates $p_{it}$. All documents with associated tokens are further investigated by LayoutLM. 

\subsection{LayoutLM for token classification} \label{subsec:tokenClass}
To make LayoutLM suitable for TCT, a linear layer is appended to the hidden states output of LayoutLM. This layer has as many neurons as potential token classes exist.

\subsection{Generation of ground truth layout data} 
\label{sec:method-ground-truth}
The key benefit of models such as LayoutLM utilizing text and layout information for VDU is their superior performance by incorporating structural properties of documents. Particularly, documents such as invoices, receipts or delivery notes are rigorously formalized complying with legal requirements. However, retrospective VDU datasets (e.g. SROIE \citep{huang2019}), particularly those being established before the publication of models such as LayoutLM, often miss this layout information in the ground truth. The latter is rather limited to token level, i.g. $l\_grossamount = 40.00 \ \textbf{€} $. In order to make these datasets available for training recent models, we have to add layout information, in particular the corresponding positions $P_i$ for the token sets $T_i$.
For this purpose we apply OCR on training images resulting in a set of tokens $T_i$, each being accompanied by image coordinates $P_i$. Our ground truth tokens $\hat{T_i}$, i.e. labeled document token labels such as grossamount or creditor name are matched to the OCR's output $T_i$ as follows:
\begin{align}
\label{eq:text-distance}
    \delta_T (T_i,\hat{T_i}) = 
        \left\{\begin{array}{lr}
        0 & \text{if } \operatorname{is\_sub}(T_i,\hat{T_i}),\\
        0 & \text{if } \operatorname{is\_sub}(\hat{T_i},T_i),\\
        \operatorname{lev(T_i,\hat{T_i})} & \text{otherwise}
        \end{array}\right.
\end{align}
The function $\operatorname{is\_sub}$ checks for the exact substring matching of OCR readings and label texts, $\operatorname{lev}$ denotes the Levenshtein distance. The substring search goes from left to right and checks if the sequence of characters of the first string, can be found in anywhere in the other given string. The Levenshtein distance is a metric for measuring the difference of two input text sequences summing the number of character alterings required for transforming one sequence into the other. An in-depth investigation can be found in \citep{yujian2007}. We've applied upper boundaries for the levenshtein distance to ensure that the searches are not taking too long.
Note that, Eq. \ref{eq:text-distance} is not suitable if exact matchings are required, e.g. in the case of currency- and date-related labels which is incorporated in our preprocessing.
If ground truth data cannot be accurately mapped to OCR readings, we omit those in our dataset preparation avoiding noisy labels for our supervised training. As a result of this step, we obtain a set of tokens $T_i$, each being accompanied by labels $\hat{t_i}$ and (layout) positions $P_i$. 

\subsection{Evaluation}
To evaluate the accuracy of the trained model with respect to the classification task the F1-Scores \eqref{eq:f1} for \textit{each token class} \citep{Geron2022} are calculated. Section \ref{sec:trainSetup} shows the classes of interest in our experiments.
To evaluate the applicability of the trained token classification model for documents in deployment, a new metric is introduced: \textit{Document Integrity Precision} (DIP). DIP is the ratio of all documents, whose tokens were completely correctly classified, divided by the count of all documents for which predictions were done.
With DIP, a connection to the business task is created during the training process. Unlike conventional metrics DIP indicates how many documents do not need further intervention when a model is deployed and thus, addresses the need for as much automation as possible. This metric is of major importance from a business perspective.

Defining a good value for DIP depends on the context in which the model is used and the wanted automation level without further interventions. The amount of documents \textit{completely correctly} classified through all token classes are denoted as $CD_t$ and the set of all classified documents is $D_t$, where $t$ denotes a subset of all documents $D$ in the dataset which are used for testing during model training. As in $CD_t$ only completely correctly documents are listed, it's important to mention that DIP in its current version does not allow partial correctness. It is rigorous in the definition of what is right since it should be used in highly automated environments in which human corrections are unwanted. Furthermore, all token classes are weighted equally for determining if a document is completely correct.

Each token can only have one correct output. However, for the case of multiple tokens being chosen with the same propability, the first one in the token list is chosen.

\begin{equation} \label{eq:recall}
    recall = \frac{TP}{TP+FN}
\end{equation}
\begin{equation} \label{eq:precision}
    precision = \frac{TP}{TP+FP}
\end{equation}
\begin{equation} \label{eq:f1}
    f1 = 2 \times \frac{precision \times recall}{precision + recall}
\end{equation}
\begin{equation} \label{eq:dip}
    DIP = \frac{CD_t}{D_t}
\end{equation}
\begin{itemize}
    \item $TP$ denotes the amount of correctly classified labels,
    \item $FP$ denotes falsely classified labels, and
    \item $FN$ denotes the amount of labels not detected by the model.     
\end{itemize}

\section{Dataset} \label{sec:dataset}
For the experiments, a private dataset is used. The dataset consists of German receipts. After applying the filtering as mentioned in \ref{sec:method-ground-truth}, the dataset had a total amount of 7456 receipts originating from 903 different creditors. Each sample consists of an image of the receipt and a JSON file, with the OCR output of the tokens with their respective coordinates in pixels. The coordinate system starts at the upper left corner of each image. Also, each token receives a token class as listed in Section \ref{sec:trainSetup}. Tokens, which do not belong to any of the mentioned classes, are labeled as \textit{None} and are not used for the calculation of DIP.  
Most creditors contribute one or few samples into the dataset as described by Figure \ref{fig:disCredFileAmount}. 12 creditors have more than 100 and up to 450 documents. For scenario S2\_100, introduced in Section \ref{sec:trainSetup}, to train the token classification model, the dataset gets divided strictly by creditors. Hence, receipts of a creditor in the train-dataset will not be included in the test-dataset and vice versa. This is done to mimic the effect of a deployed model in production needs to evaluate invoices of an unknown entity, f.e. new vendors.     
\begin{figure}
    \centering
    \includegraphics[width=0.7\columnwidth]{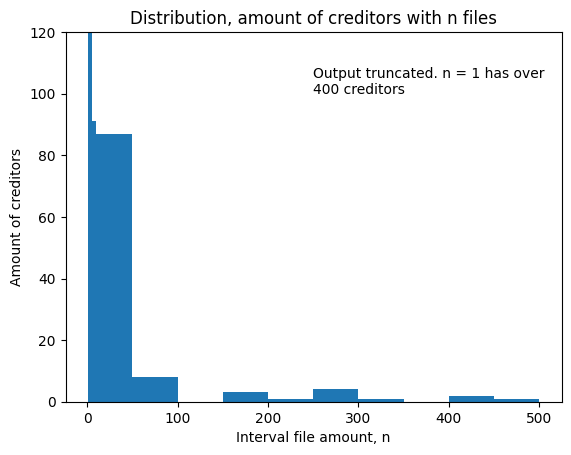}
    \caption{Distribution fileamount of creditors. For up to 10 files, the histogram has a range per bin of 1, from 10-50 files of 10 and afterwards 50.}
    \label{fig:disCredFileAmount}
\end{figure}

\section{Training Setup / Experiments} \label{sec:trainSetup}
For all our experiments, the underlying model, applied hyperparameters and hardware specifications are listed in Table~\ref{tab:hypParHardw}. Classes/Labels focused for our experiments are:

\begin{table*}[ht]
\centering
\caption{Model, hyperparameters used for the training and hardware specifications.}
\label{tab:hypParHardw}
\begin{tabular}{@{}ll@{}}
\toprule
\textbf{Model} & \textbf{Value}                    \\ 
\midrule
Model Architecture      & LayoutLM                          \\
Model Name              & microsoft/layoutlm-base-uncased   \\
Tokenizer Name          & microsoft/layoutlm-base-uncased   \\
\midrule
\textbf{Hyperparameter} & \textbf{Value}                    \\ 
\midrule
Learning Rate           & 0.001                             \\
Percentage Train        & 0.8                               \\
Batch Size Train        & 8                                 \\
Batch Size Test         & 4                                 \\
Epochs                  & 200                               \\
Optimizer               & SGD                               \\
\midrule
\textbf{Hardware}       &                                   \\ 
\midrule
Kernel                      & Linux-5.19.0-45-generic-x86\_64-with-glibc2.31 \\
GPU                     & NVIDIA RTX A4000                  \\
RAM                     & 45GB                               \\
\bottomrule
\end{tabular}
\end{table*}

\begin{table*}[ht]
\centering
\caption{F1-Scores and DIPs for each label in training scenario S1\_100 and S2\_100 (see Section \ref{sec:dataset})}
\label{tab:s1_training_runs}
\begin{tabularx}{\textwidth}{X|X|X|X|X|X|X}
\toprule
\textbf{Scenario} & \tiny{\textbf{invoicenumber}} & \tiny{\textbf{documentdate}} & \tiny{\textbf{creditorname}} & \tiny{\textbf{grossamount}} & \tiny{\textbf{netamount}} & \tiny{\textbf{DIP}}\\
\midrule
S1\_100                
  & 0.986                                              
  & 0.995                                
  & 0.976                                
  & 0.982                                
  & 0.926                                
  & \textbf{0.796} \\
S2\_100                
  & 0.923                                            
  & 0.983                                
  & 0.617                                
  & 0.927                                
  & 0.785                                
  & \textbf{0.225} \\ \bottomrule
\end{tabularx}
\end{table*}

\begin{itemize}
    \item document date                                
    \item invoice number                               
    \item creditor name                                
    \item net amount                                   
    \item gross amount                                 
\end{itemize}
\section{Results} \label{sec:res}
Results of both training scenarios can be found in Table~\ref{tab:s1_training_runs} and Figure \ref{fig:comparisonf1dip}. 

\section{Discussion} \label{sec:discussion}
After our experiments, we've created two models trained with the TCT task. One for each Scenario. We've calculated the F1-Score for each token class of interest and the DIP for each respective scenario. The concrete values of our experiments are listed in Table \ref{tab:s1_training_runs}. A visual representation but with the avg. F1-Scores for each scenario can be found in Figure \ref{fig:comparisonf1dip}.

In run S1\_100, it is observable that the model generated high F1-Scores for each token class. The avg. value is round about 0.98. One could say that this model is well suited since it classified the tokens correctly through all token classes with high confidence. Hence, this model could be used in production by only looking at the F1-Scores. On the other hand, the DIP has a score of round about 0.8. This means that in inference probably 20 of 100 documents need to be manually corrected.

In run S2\_100, the discrepancy between DIP and F1-Score as representative of conventional classification metrics is even more highlighted. While having an avg. F1-Score of round about 0.8, which is not optimal but not bad either, depending on the requirements, DIP deteriorates to round about a quarter of the F1-Score. Therefore, it can be assumed that 80 of 100 documents will need manual post-processing steps to correct the mistaken classifications. 

It is important to note that the results of label 'netamount' suffered due to challenges in ground-truth mapping, mentioned in \ref{sec:methodology}. However, we assume that this does not entail substantial problems in regards of the validity of the presented results. With a perfect mapping of ground truth data, we expect better results for the conventional classification metrics as well as DIP. However, we assume that the relations between the changes in the conventional metrics and DIP remain similar. To visualize the results, Figure \ref{fig:resultVisualization} shows three documents and their respective predictions for the labels defined in Section \ref{sec:dataset}, created by the model trained in scenario S1\_100. The pictures focus on wrong predictions for one specific label. 

\begin{figure}
    \centering
    \includegraphics[width=\columnwidth]{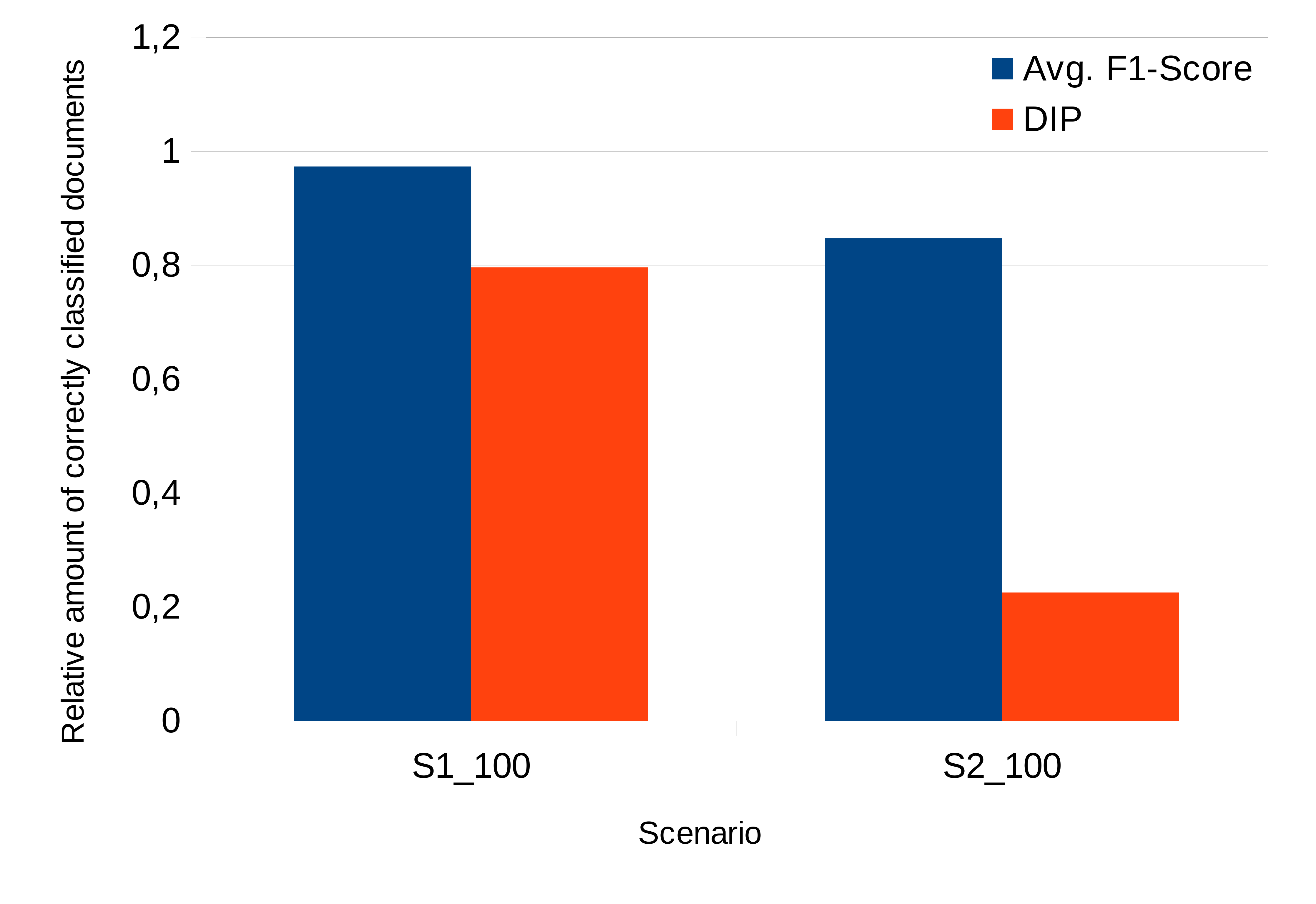}
    \caption{Comparison of the avg. F1-Scores and DIP for each scenario.}
    \label{fig:comparisonf1dip}
\end{figure}

\begin{figure*}[h]
    \centering
    \begin{subfigure}{0.3\textwidth}
        \centering
        \includegraphics[width=\textwidth]{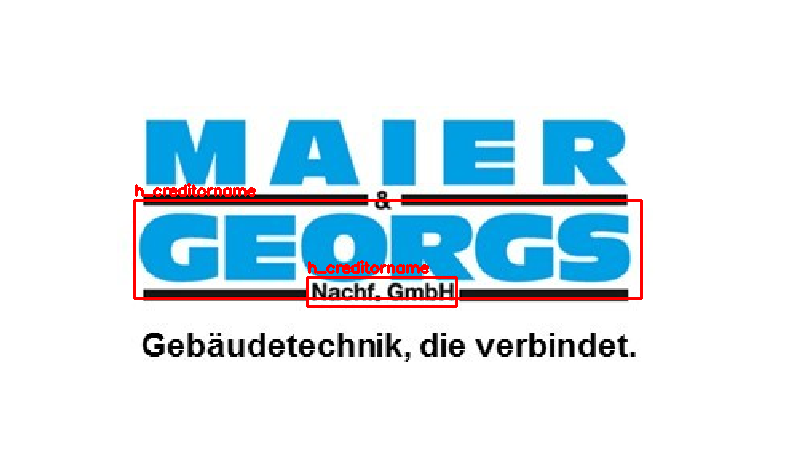}
        \caption{First part of the creditorname was not recognized.}
    \end{subfigure}
    \hfill
    \begin{subfigure}{0.3\textwidth}
        \centering
        \includegraphics[width=0.8\textwidth]{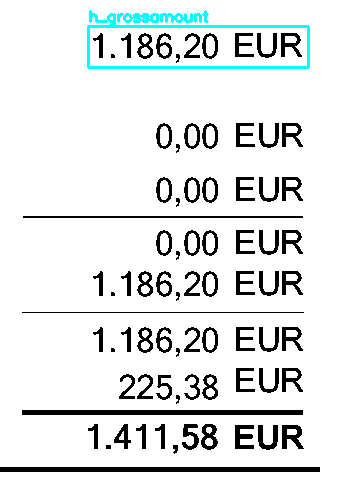}
        \caption{The grossamount is 1.411,58 (In German internalization). However, the Model predicted 1.186,20.}
    \end{subfigure}
    \hfill
    \begin{subfigure}{0.3\textwidth}
        \centering
        \includegraphics[width=\textwidth]{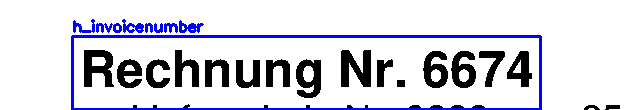}
        \caption{The whole expression before the invoicenumber (in German: Rechnungsnummer, abbrev.: Rechnung Nr.) was also classified as it.}
    \end{subfigure}
    \caption{\textit{Failure cases}: Extracts from sample invoices of run S1\_100 whose predictions for label creditorname, grossamount and invoicenumber are respectively wrong. DIP indicates that 20\% of all documents need to be manually corrected. However, in terms of individual labels, this effect is averaged out, resulting in overall high F1 scores, which leads to a falsely very good assessment of the prediction quality.}
    \label{fig:resultVisualization}
\end{figure*}

Our results imply that conventional classification metrics are not enough to decide whether a model should be deployed or not. Further metrics are needed to decide if a model is suitable for deployment, regarding the business task. For the TCT task for VDU in environments with the desire of high automation magnitudes DIP can assist in the decision of suitability. The results further imply that developers of deep learning models need to pay attention to their business task and cannot only rely on conventional classification metrics to define their model performance in production. If new models or model-architectures are invented, researchers and developers need to invent new metrics which respect the aspect of applicability of a model in inference.

To the best of our knowledge this work is the first of its kind. The applicability of model and ways to define them are not considered in research. 

However, there are limitations. First of all, this work did not validate the results of DIP in inference. Furthermore, DIP is rigorous in its definition of correctness. Only completely correct documents are considered as that. Therefore, the use-cases of DIP are more restricted than for more dynamic metrics, which allow partial correctness.

\section{Conclusion} \label{sec:concl}

This paper demonstrated the need to incorporate further classification metrics to evaluate the applicability of a classification model for businesses, specifically a model for Token Classification Task (TCT) in documents. To address this problem, Document Integrity Precision (DIP) was introduced as an evaluation metric. DIP demonstrates how many classified documents still need manual interventions before the results can be processed further. Furthermore, a visual document analysis system was trained on several training data sizes with a constant test dataset to understand changes in model accuracy with fewer data in relation to DIP. Although the differences in conventional classification metrics are small, DIP halves, indicating a far less suitable model for deployment.

DIP shifts the focus to the applicability and the business task of document understanding. It also highlights the discrepancy in performance during model inference as a result of misleading conventional training metrics. By revealing this discrepancy, DIP can help avoid deploying models whose outputs potentially require a high amount of manual labor to be correct and usable. This is important because the experiments in this paper show that conventional metrics such as recall, precision, and F1-Score can be high, indicating that the model is predicting the correct labels. This can lead to a false sense of certainty about the model's performance in inference, while in reality, manual corrections are still needed, occupying labor and slowing down the automation process.

Further research should evaluate DIP's prediction rate of manual labor needed and compare it to the real amount occurring in a production system. Based on these results, DIP could be further refined. One question is how to modify DIP to support partial correctness. In other words, define some of the token classes of interest as optional. Our current implementation is mainly viable for business environments with a desire for high automation without further manual post-processing. This currently limits the potential business cases in which DIP can be used. Furthermore, additional research areas could include how DIP could or could not be used in hybrid evaluation approaches.

However, the experiments show that DIP is currently fundamentally important in assessing the applicability of a model for document understanding in a business context. It highlights the need for further evaluation metrics focused on the business task to make informed decisions about whether a model should be deployed or not. However, it does so with a focus on the TCT for Visual Document Understanding (VDU). Further research also needs to create novel metrics for other training tasks and business cases, such as regression or prediction.

\printbibliography
\end{document}